\documentclass[letterpaper]{article}

\usepackage{natbib,alifeconf}  
\usepackage{url,hyperref}
\usepackage{booktabs}
\usepackage{amsmath} 
\usepackage{amssymb}
\usepackage{amsfonts}
\usepackage{lipsum}
\usepackage{cleveref} 
\usepackage{dblfloatfix}
\usepackage{cuted}
\usepackage{caption}
\usepackage{xcolor}

\newcommand{\bh}[1]{{\color{black}#1}}  
\usepackage[normalem]{ulem}                

\newcommand\blfootnote[1]{%
  \begingroup
  \renewcommand\thefootnote{}\footnote{#1}%
  \addtocounter{footnote}{-1}%
  \endgroup
}

\usepackage[rightcaption]{sidecap}

\title{BraiNCA: brain-inspired neural cellular automata and applications to morphogenesis and motor control}

\author{
    Léo Pio-Lopez$^{1}$,
    Benedikt Hartl$^{1}$,
    Michael Levin$^{1, 2}$ \\
    \mbox{}\\
    $^1$Allen Discovery Center at Tufts University, Medford, MA, USA \\
    $^2$Wyss Institute for Biologically Inspired Engineering at Harvard University, Boston, MA, USA\\
    Corresponding Author: \href{mailto:michael.levin@tufts.edu}{michael.levin@tufts.edu}
} 

\begin{document}

\maketitle

\begin{abstract}
Most of the Neural Cellular Automata (NCAs) defined in the literature have a common theme: they are based on regular grids with a Moore neighborhood (one-hop neighbour). They don't take into account long-range connections and more complex topologies as we can find in the brain. In this paper, we introduce BraiNCA, a brain-inspired NCA with an attention layer, long-range connections and complex topology. BraiNCAs shows better results in terms of robustness and speed of learning on the two tasks compared to Vanilla NCAs establishing that incorporating attention-based message selection together with explicit long-range edges can yield more sample-efficient and damage-tolerant self-organization than purely local, grid-based update rules. These results support the hypothesis that, for tasks requiring distributed coordination over extended spatial and temporal scales, the choice of interaction topology and the ability to dynamically route information will impact the robustness and speed of learning of an NCA. More broadly, BraiNCA provides brain-inspired NCA formulation that preserves the decentralized “local update” principle while better reflecting non-local connectivity patterns, making it a promising substrate for studying collective computation under biologically-realistic network structure and evolving cognitive substrates.
\end{abstract}

Submission type: \textbf{Full Paper}\\

Data/Code available at: \url{https://github.com/LPioL/BraiNCA}
\blfootnote{\textcopyright Published under a Creative Commons Attribution 4.0 International (CC BY 4.0) license.}

\section{Introduction}

NCAs extend classical cellular automata by introducing learnable, adaptive local rules, typically implemented by Artificial Neural Networks (ANNs).
They are spatially distributed computational systems where individual cells maintain internal states and coordinate through message passing with their neighbors \cite{mordvintsev_growing_2020, hartl2025neural}. The collective decision-making can give rise to system-level outcomes and NCAs excel at modeling tasks like development and regeneration \cite{mordvintsev_growing_2020, pio2023scaling, hartl2024evolutionary, guichard2025engramnca, 10.1145/3638529.3654150, pande2023hierarchical}, aging \cite{pio2025aging, cavuoti2022adversarial}, but can also be efficiently used as robotic controllers \cite{hartl2025neuroevolution, horibe2022severe, horibe2021regenerating}. Recently, they even have shown competitive results at abstraction and reasoning tasks on challenging benchmarks like the ARC corpus \cite{guichard2025arcnca, xu2025neural}. Overall, this architecture makes the bridge between biological principles based on the multi-competency architecture \cite{levin2022technological, levin2023darwin, fields2022competency, dodig2022cognition, ermentrout1993cellular}, and modern artificial intelligence \cite{hartl2025neural}.

\begin{figure}
    \centering
    \includegraphics[width=\linewidth]{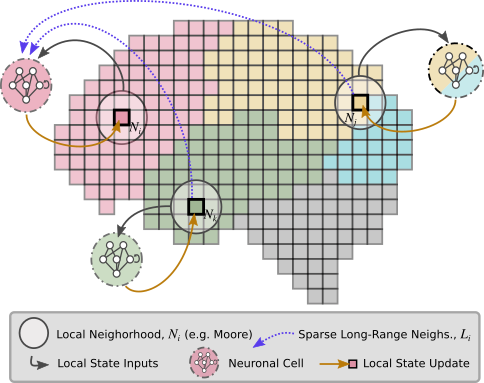}
    \caption{\textbf{Schematic illustration of the BraiNCA model.} Every cell $i$ not only integrates its local neighborhood $\mathcal{N}_i$ via attention, but also long-range signals from sparsely connected distant cells $j,k\in\mathcal{L}_i$ for recurrent state updates. Flexible topologies are allowed, departing from NCAs with regular, e.g., square grid-layouts.}
    \label{fig:model}
\end{figure}

Historically, NCAs are a modern extension of classical cellular automata (CA) \cite{kari2005theory} in which the discrete, hand-designed local transition rule is replaced by a learnable update function, typically implemented as a small ANN shared across all cells. Like standard cellular automata, an NCA consists of many simple units arranged on a grid that update their states over discrete time steps using only local information, yet can collectively generate rich global dynamics. 
The key difference is that the update rule is optimized from data (e.g., by gradient-based learning or evolution), allowing the system to acquire adaptive local interaction policies that realize a desired macroscopic behavior such as self-assembly, morphogenesis, or robust pattern maintenance. More formally, each cell maintains a  vector-valued state and updates it by applying the same parameterized function to its own state and the states of its neighbors, commonly in a Moore neighborhood on a 2D lattice \cite{hartl2025neural}.  In widely used NCA formulations, the cell state is partitioned into visible channels (for instance, RGB plus an opacity/liveness channel) and additional hidden channels that support distributed computation, memory, and coordination during iterative development from a small seed. As described above, this paradigm has shown impressive capabilities on tasks requiring robustness and generalization in open-ended, dynamic environments, biological applications and many other sub-domains of machine learning \cite{hartl2025neural, guichard2025arcnca, xu2025neural}.


Most of the NCA defined in the literature have a common theme: they are based on regular grids with a Moore neighborhood (one-hop neighbour). They don't take into account long-range connections and more complex topologies as we can find in the brain. Accordingly, recent extensions of CA and NCAs replace the lattice with an arbitrary interaction graph, where each node updates from its adjacency-defined neighborhood using permutation-invariant aggregation in the spirit of graph neural networks, thereby supporting non-Euclidean substrates and heterogeneous connectivity, e.g. Graph-CA, Graph-NCA and GNCAs \cite{grattarola2021learning, gala2023n, moyo2025self}.  This graph formulation can naturally express long-range connections (via explicit edges) and complex topologies, while retaining the same decentralized “local rule, global behavior” principle that underlies grid-based/Vanilla NCAs.  Later work extended GNCA to  E(n)-equivariant GNCA formulations \cite{gala2023n}, supporting more faithful interaction modeling on continuous or spatially embedded graphs. Complementary approaches incorporate a self-organizing GNCA with the interaction graph itself adaptive by learning that can reorganize edges over time, so that information flow can change as a function of the evolving node states and thereby support history-dependent computation \cite{moyo2025self}. Another study focused on growing the graph structure itself from a seed, effectively treating morphogenesis as the construction of a computational substrate over developmental time \cite{barandiaran2025growing}. While GNCAs introduces a powerful extension to NCAs by formulating them with a graph approach, they have not been compared to Vanilla NCA on biological and motor tasks nor have explored different long-range connectivity patterns to resolve these tasks.


The proponents of brain-inspired AI  argue that progress will come not only from more data or computational resources but from embedding neurobiological constraints on connectivity and organization into model design (e.g., sparse, structured, and physically grounded wiring rather than unconstrained dense graphs) \cite{hassabis2017neuroscience, li2024brain}. Brain-inspired AI seeks to incorporate organizing principles of nervous systems—where computation emerges from constrained connectivity, distributed dynamics, and continual adaptation—rather than relying solely on generic dense architectures trained end-to-end.  A recurring principle is wiring economy, which frames brain networks as optimizing function under constraints on connection length and material/energy cost \cite{bullmore2012economy}, motivating machine architectures that explicitly regularize or constrain connectivity instead of assuming all-to-all communication.  Related work in network science formalizes how sparse long-range “shortcuts” can coexist with strong local clustering (small-world structure), improving global integration without sacrificing local specialization \cite{watts1998collective}, and connectome studies further highlight hub-like “rich-club” cores that support efficient communication across modules \cite{van2011rich}.  Another core principle is topographic organization—classically modeled by self-organizing maps that preserve neighborhood relations in learned representations \cite{kohonen1982self}—and grounded in somatotopic organization observed in human cortex \cite{penfield1937somatic}, together motivating AI architectures that embed spatial/topographic inductive biases when mapping sensory or body-related variables.  For learning and inference, predictive-processing theories propose hierarchical generative models in which feedback conveys predictions and feedforward pathways transmit residual errors \cite{rao1999predictive}, and the free-energy/predictive-coding framework connects these error signals to learning and perceptual phenomena at the cortical level \cite{friston2016activeinference}.  

In this paper, we introduce BraiNCA, a brain-inspired NCA with an attention layer, long-range connections and non-grid-based topologies \bh{, see Fig.~\ref{fig:model}}. Building on the GNCA paradigm of learned,  message passing over arbitrary graphs, BraiNCA replaces purely local aggregation with attention-weighted neighborhood integration, allowing each node to dynamically select and gate information from its neighbors as a function of context. In addition, we explicitly model long-range interactions by augmenting the underlying connectivity (e.g., via multi-hop edges or explicit long-range connections), thereby enabling information flow beyond one-hop neighborhoods as in Vanilla NCAs and better matching the heterogeneous, non-local wiring patterns characteristic of brain networks. Concretely, BraiNCA maintains node-wise hidden channels to store latent state and update history, while the attention mechanism modulates both short- and long-range messages to support flexible, state-dependent coordination across the graph. We tested this architecture on a morphogenetic task and a motor task (lunar lander). BraiNCAs shows better results in terms of speed of learning for morphogenesis, and in robustness and speed of learning for the lunar lander compared to Vanilla NCAs. This establishes that incorporating attention-based message selection together with explicit long-range connectivity and new topologies can yield more sample-efficient and robust self-organization than purely local, grid-based update rules.  More broadly, BraiNCA provides a brain-inspired NCA formulation that preserves the decentralized “local update” principle while better reflecting non-local connectivity patterns, making it a promising substrate for studying collective computation under biologically-realistic network structure and self-evolving cognitive substrates. 


\section{BraiNCA Model}

\bh{BraiNCA supercharges the purely self-organizied state-update dynamics of an NCA with a flexible integration of both local and long-range neighborhoods via attention and message-passing mechanisms.
While short-range connections integrate information locally, scale-free hub connections can bypass communication globally.}

\bh{
A BraiNCA consists of $N$ localized cells that are distributed on---but not restricted to---an $L\times M$ regular grid.
Each cell $i\in\{1,\dotsc,N\}$ maintains a $C$-dimensional state $\mathbf{c}_i^t\in\mathbb{R}^{C}$ across contiguous timesteps $t$, and might optionally be conditioned by environmental input $\rho_i^t$.
Formally, the extended state $\mathbf{s}_i^t=[\mathbf{c}_i^t;\rho_i^t]\in\mathbb{R}^S$ concatenates cell state $\mathbf{c}_i^t$ and conditional input $\rho_i^t$.
The local neighborhood of cell $i$ is $\mathcal{N}_i=\{i_\nu\}_{\nu=0}^{N_i}$, defining a set of $N_i$ neighboring cell states $\{\mathbf{c}_{k}^t\}_{k\in\mathcal{N}_i}$; $\nu=1,\dotsc,N_i$ iterates over the local neighbors of cell $i$ (typically---but not restricted to---a Moore neighborhood), $k=i_\nu$ represents the global grid index of cell $i$'s neighbor $i_\nu$, and $\nu=0$ expresses cell identity $i\equiv i_{0}$.
Analogously, every cell $i$ aggregates state-information $\{\mathbf{c}_{k^\prime}^t\}_{k^\prime\in\mathcal{L}_i}$ from distant neighbors $k^\prime=i_\mu$ from a sparse long-range neighborhood $\mathcal{L}_i=\{i_\mu\}_{\mu=0}^{L_i}$, see Fig.\ref{fig:model}.
}

\begin{figure}[t]
    \centering
    \includegraphics[width=\linewidth]{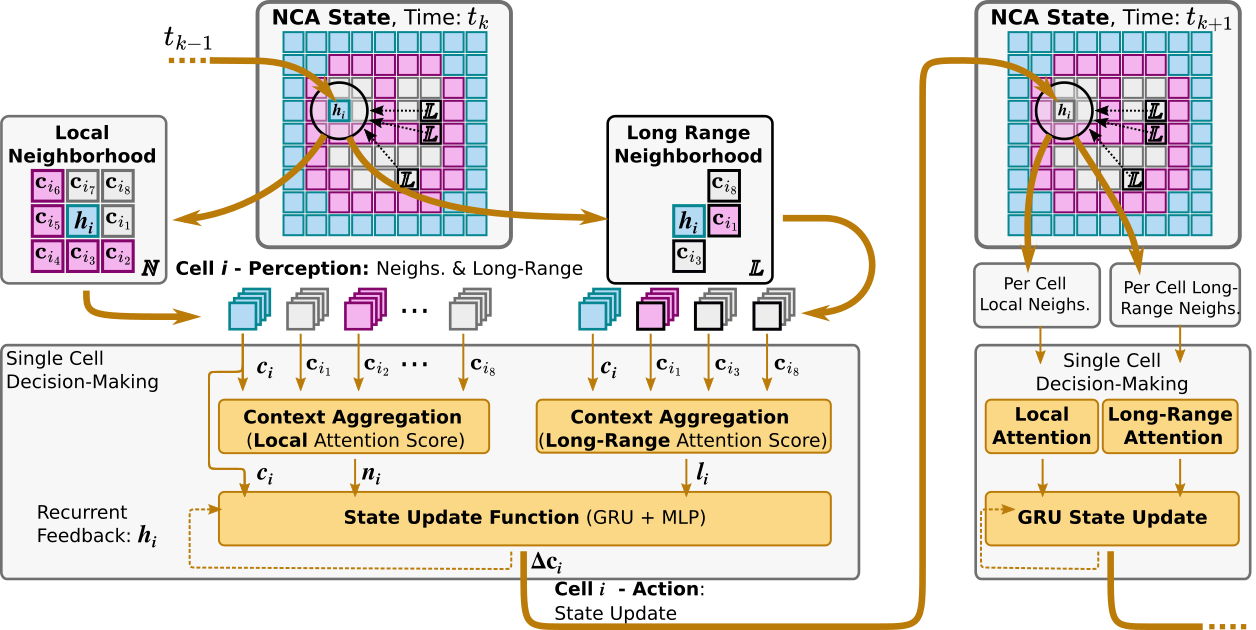}
    \caption{\textbf{Flow-diagram of the BraiNCA architecture.} Each cell’s state is updated by aggregating information from both local and long-range neighborhoods through attention-based context encoding. The combined signals are processed via a recurrent update function (GRU + MLP), enabling spatially distributed coordination in morphogenesis experiments and sensory-motor control of RL agents.}
    \label{fig:architecture}
\end{figure}

\bh{
Cell-state updates are formally given by
\begin{equation}
    \mathbf{c}_i^{t+1}=f_\theta\left(\{\mathbf{c}_{k}^t\}_{k\in\mathcal{N}_i}, \{\mathbf{c}_{k^\prime}^t\}_{k^\prime\in\mathcal{L}_i}, \mathbf{u}_{t}, \mathbf{o}_t\right)
    \label{eq:update}
\end{equation}
and contextualize cell-states over time $t$ by aggregating local $\{\mathbf{c}_{k}^t\}_{k\in\mathcal{N}_i}$, long-range $\{\mathbf{c}_{k^\prime}^t\}_{k^\prime\in\mathcal{L}_i}$, and (optionally) system-level feedbak $\mathbf{u}_t$ and external information $\mathbf{o}_t$ via an update function $f_\theta(\cdot)$, i.e., an ANN parametrized with parameters $\theta$.
Both local and long-range neighborhoods are aggregated via respective attention mechanisms before being jointly processed by the state update module that employs a Gated Recurrent Unit (GRU), see Fig.~\ref{fig:architecture}.
}

\bh{
For each cell $i$ we compute the aggregated local, $\mathbf{n}_i^t$, and long-range neighborhoods, $\mathbf{l}_i^t$, at every timestep $t$ via
\begin{eqnarray}
a_{ij}^t &= f_{\text{attn}}^{(\mathcal{N})}([\mathbf{s}_i^t;\mathbf{s}_j^t]), \quad
b_{ij}^t &= f_{\text{attn}}^{(\mathcal{L})}([\mathbf{s}_i^t;\mathbf{s}_j^t]), \label{eq:state:attention:map}\\
\alpha_{ij}^t &= \frac{\exp(a_{ij}^t)}{\sum_{k \in \mathcal{N}_i}\exp(a_{ik}^t)}, \quad
\beta_{ij}^t &= \frac{\exp(b_{ij}^t)}{\sum_{k^\prime \in \mathcal{L}_i}\exp(b_{ik^\prime}^t)}, \label{eq:state:attention:normalized} \\
\mathbf{n}_i^t &= \sum_{k \in \mathcal{N}_i} \alpha_{ik}^t \mathbf{h}_k^t, \quad
\mathbf{l}_i^t &= \sum_{k^\prime \in \mathcal{L}_i} \beta_{ik^\prime}^t \mathbf{h}_{k^\prime}^t, \label{eq:state:attention:aggregation}
\end{eqnarray}
where $f_{\text{attn}}^{(\mathcal{N})}(\cdot),f_{\text{attn}}^{(\mathcal{L})}(\cdot):\mathbb{R}^{2S}\rightarrow \mathbb{R}^A$ are multilayer perceptron networks (MLPs) that operate on concatenated extended state pairs $[\mathbf{s}_i^t; \mathbf{s}_j^t]\in\mathbb{R}^{2S}$ for local, $\mathcal{N}_i$, and long-range neighbors, $\mathcal{L}_i$, respectively; 
$a_{ij}^t$ and $b_{ij}^t\in\mathbb{R}^A$ are pair-wise attention scores,
and $\alpha_{ij}^t$ and $\beta_{ij}^t$ are the normalized attention weights for local and long-range neighbors, respecivelty.
}

\bh{
The extended state information $\mathbf{s}_i^t$ along-side the aggregated local and long-range neighborhoods $\mathbf{n}_i^t$ and $\mathbf{l}_i^t$ are first combined into an interaction vector $\mathbf{z}_i^t$ before being converted into the input $\mathbf{m}_i^t$ for the downstream update function:
\begin{equation}
    \mathbf{z}_i^t = f_Z(\mathbf{s}_i^t, \mathbf{n}_i^t, \mathbf{l}_i^t),\quad
    \mathbf{m}_i^t = f_{\textnormal{msg}}(\mathbf{z}_i^t; \mathbf{u}_t, \mathbf{o}_t).
    \label{eq:state:composition}
\end{equation}
$f_Z(\cdot)$ denotes a generic composition operator that combines  $\mathbf{s}_i^t$, $\mathbf{n}_i^t$, and $\mathbf{l}_i^t$ into $\mathbf{z}_i^t$; the specific form of $f_Z(\cdot)$ is deliberately left unspecified to maintain generality, but will be defined in the examples below.
$f_{\textnormal{msg}}(\cdot)$ is an MLP that processes the interaction vector $\mathbf{z}_i^t$, and optionally puts it in context of system-level feedback $\mathbf{u}_t$ and external input $\mathbf{o}_t$.  
While $\mathbf{s}_i^t$ may act as a residual, long-range integration is achieved via concatenation or additive combination of $\mathbf{n}_i^t$ and $\mathbf{l}_i^t$, for instance. 
In contrast, a purely local (Vanilla) NCA is recovered by omitting long-range contributions.
}

\bh{
The state update mechanism employs a GRU followed by a feedforward refinement network, $f_{\textnormal{refine}}$, with refinement connection to yield $\mathbf{c}_i^{t}\mapsto\mathbf{c}_i^{t+1}$:
\begin{eqnarray}
    \mathbf{h}_i^t &=& f_{\textnormal{GRU}}(\mathbf{m}_i^t, \mathbf{h}_i^{t-1}),\quad
    \mathbf{r}_i^t = f_{\textnormal{refine}}(\mathbf{h}_i^t)\\
    \mathbf{c}_i^{t+1} &=& \mathbf{h}_i^t + \mathbf{r}_i^t.
    \label{eq:state:update:GRU}
\end{eqnarray}
$f_{\textnormal{GRU}}(\cdot)$ represents the GRU which operates on a message input $\mathbf{m}_i^t\in\mathbb{R}^M$ to maintain a hidden state $\mathbf{h}_i^t\in\mathbb{R}^C$ across time $t$, $\mathbf{r}_i^t\in\mathbb{R}^C$ is the refinement update, and $\mathbf{c}_i^{t+1}\in\mathbb{R}^C$ is the updated cell state.
}


\section{\bh{Experimental Setup}}
\subsection{Task 1: Morphogenesis}

\subsubsection{Problem Formulation}

In the morphogenesis task, a population of $N = 256$ cells arranged on a 16×16 grid must collectively organize from random initial states into a target spatial pattern. The target pattern is a smiley face consisting of three discrete cell types: background cells, face boundary cells, and facial feature cells\bh{, c.f., Fig.~\ref{fig:architecture}}. We chose it as is it similar to the bioelectrical prepattern we can observe during frog embryogenesis \cite{vandenberg2011v} and establishing this pattern is key in (frog) development. During simulations, each cell must autonomously determine its type based solely on interactions with neighboring cells, without access to global positional information or external supervision during the self-organization process.

The task 
\bh{contrasts}
the system's ability to achieve robust pattern formation through purely local interactions \bh{(Vanilla NCA) and long-range connectivity (long-range BraiNCA)}. Success is defined as reaching 98\% or greater pixel-wise accuracy in matching the target pattern, measured after a fixed developmental period of \bh{$T=35$}
cellular update steps. Training consists of repeated developmental episodes, each beginning from a random initialization, with network parameters updated via gradient descent to minimize the discrepancy between achieved and target patterns.

\subsubsection{Architecture Specification}

For morphogenesis, each cell maintains a 
state \bh{$\mathbf{c}_i^t=(c_{i,0}^t, \dotsc, c_{i,C-1}^t)$ with $C=9$; no other input is used, $\mathbf{u}^t_i=\mathbf{o}^t=\varnothing$, and $\mathbf{c}_i^t\equiv\mathbf{s}_i^t$}. The first \textit{three} dimensions encode a probability distribution over the three cell types via softmax transformation, serving as the cell's visible  phenotype. The remaining 
\bh{\textit{six}}
dimensions constitute the hidden states\bh{, i.e.,} 
latent memory or communication channels used for coordination and computation with 
\bh{any}
connected cells\bh{; in contrast to the visible states, hidden states do not contribute to the loss directly.} 

\begin{figure} 
\centering
    \includegraphics[width=1\linewidth]{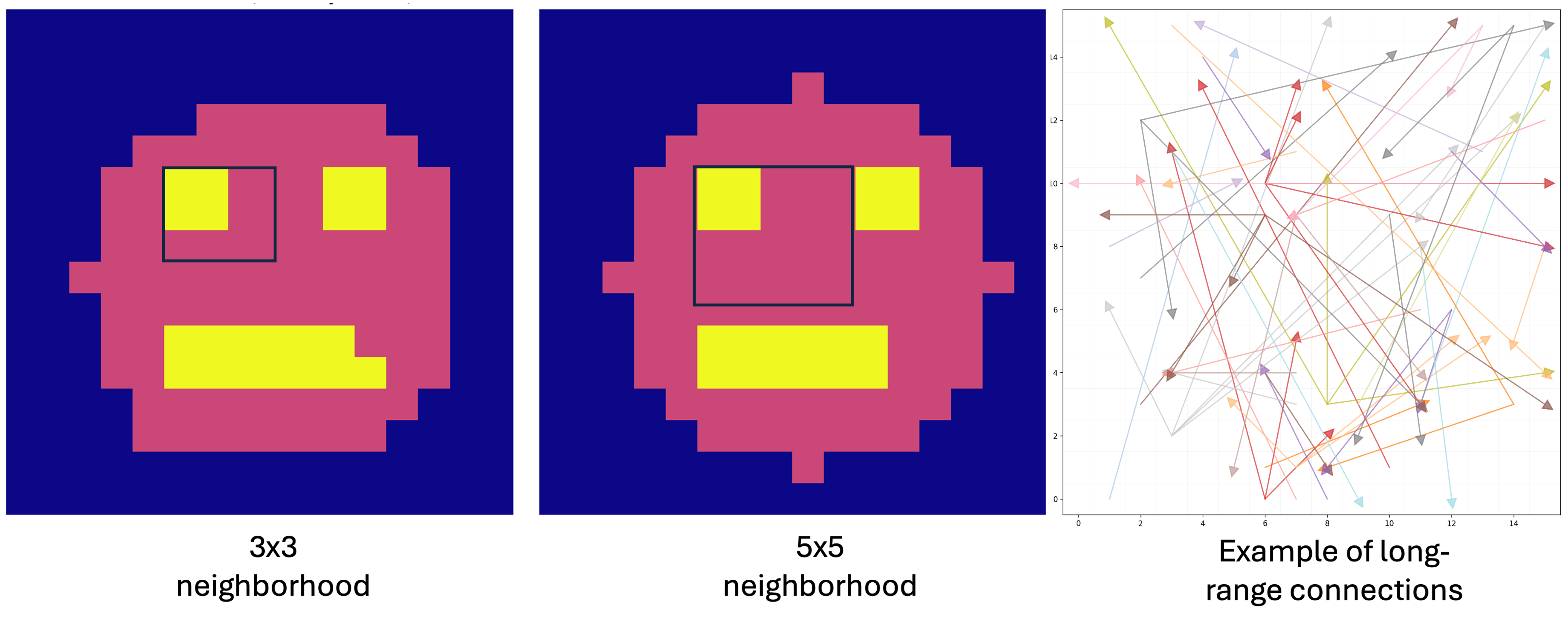}
    \caption{\textbf{Morphogenesis conditions}. Left: target pattern and cell labeling for the 3×3 neighborhood condition. Middle: target pattern for the 5×5 neighborhood condition. Right: example of long-range connections used in long-range variants. We made it sparse to mimic scale-free networks connectivity as we can find in the brain.}
    \label{fig:results_morpho}
\end{figure}

\bh{For both Vanilla and long-range conditions, we choose a local Moore neighborhood $\mathcal{N}$. 
For the BraiNCA, long-range connections $\mathcal{L}$ are initialized with task-based constraints. Long-range connectivity is generated using a scale-free-inspired random procedure, with Zipf-distributed hub degree capped at a maximum of six outgoing connections per hub. Both local and long-range attention maps $\mathbf{a}_{ij}^t$ and $\mathbf{b_{ij}^t}$, Eq.~(\ref{eq:state:attention:map}), use a two-layer MLP (64 hidden units, GELU), respectively.
For the Vanilla NCA, long-range inputs are entirely dismissed in the interaction vector $\mathbf{z}_i^t=[\mathbf{c}_i^t;\mathbf{n}_i^t]$, c.f., Eq.~(\ref{eq:state:composition}), while for the long-range case we use $\mathbf{z}_i^t=[\mathbf{c}_i^t;\mathbf{n}_i^t+\mathbf{l}_i^t]$;
here, $\mathbf{n}_i^t$ and $\mathbf{l}_i^t$ are the attention-aggregated local and long-range neighborhoods defined in Eq.~(\ref{eq:state:attention:aggregation}).}

\bh{For both Vanilla and long-range conditions, the state update mechanism $\mathbf{c}_i^t\mapsto \mathbf{c}_i^{t+1}$ is given by Eq.~(\ref{eq:state:update:GRU}), but using their respective interaction vectors $\mathbf{z}_i^t$.
The GRU $f_\textnormal{GRU}(\cdot)$ has $C=9$ hidden units and the refinement MLP $f_{\text{refine}}(\cdot)$ has two dense layers ($C=9$ units, intermediate GELU).}
%
%
%
%
%


The visible phenotype $\hat{s}_i^t$ is decoded from the first three channels of the hidden state,
\bh{$\mathbf{c}_{i,\mathrm{vis}}^t=({c}_{i0}^t, {c}_{i1}^t, {c}_{i2}^t)$ as
\begin{equation}
    \hat{s}_i^t=\arg\max_{k\in\{0,1,2\}} \mathbf{c}_{i,\mathrm{vis}}^t[k], 
\end{equation}
and the cell-type probability $p(s_i^t = k)$ for $k\in\{0,1,2\}$ is
\begin{equation}
p(s_i^t = k)=\frac{\exp(\mathbf{c}_{i,\mathrm{vis}}^t[k])}
{\sum_{k'=0}^{2}\exp(\mathbf{c}_{i,\mathrm{vis}}^t[k'])}.
\end{equation}}
%


\subsubsection{Training Procedure}

Training employs supervised learning with cross-entropy loss\bh{, $\mathcal{L}_{\text{morph}}$,} computed over all visible cell types at the final timestep of each episode:
\begin{equation}
\mathcal{L}_{\text{morph}} = -\frac{1}{N} \sum_{i=1}^{N} \sum_{k=0}^{2} \mathbf{1}[s_i^{\text{target}} = k] \log p(s_i^T = k)
\end{equation}

where $T = 35$ denotes the final timestep and $\mathbf{1}[\cdot]$ is the indicator function, i.e., $\mathbf{1}[A]=1$ if condition $A$ is true and $0$ otherwise (so only the target class contributes at each cell). Each training episode initializes all cell states by sampling from a Gaussian distribution: 
$\mathbf{c}_i^0 \sim \mathcal{N}(0, 0.1 \cdot \mathbf{I})$, 
where $\mathbf{I}$ is the $C \times C$ identity matr$T=35$ update steps, after which the loss is computed and parameters updated via backpropagation through time using the Adam optimizer with learning rate $\alpha = 10^{-3}$, $\beta_1 = 0.9$, and $\beta_2 = 0.999$.

Training continues for up to 5,000 episodes per run, with early stopping when accuracy reaches or exceeds the 98\% threshold. Network parameters (weights and biases in attention networks, message network, GRU, and refinement network) are initialized using Xavier uniform initialization for weights and zeros for biases.

\subsubsection{Experimental Conditions}

We evaluated four architectural variants: local neighborhood size (3×3 Moore neighborhood providing 8 neighbors per cell, versus 5×5 Moore neighborhood providing 24 neighbors) and connectivity structure (Vanilla with only local connections, versus long-range augmented with additional distant connections as described in Section 2.1.3). This yields four conditions: 3×3 Vanilla, 3×3 + Long-Range, 5×5 Vanilla, and 5×5 + Long-Range connections. Each condition was trained for 50 independent runs using random seeds 42 through 91.

\subsection{Task 2: Lunar Lander}

\subsubsection{Problem Formulation}

The Lunar Lander task requires learning a control policy for landing a spacecraft on a designated landing pad using discrete thruster commands. The environment, provided by Gymnasium LunarLander-v3, presents a  control problem where an agent receives 8-dimensional observations \bh{$\mathbf{o}^t\in\mathbb{R}^8$} (horizontal and vertical position, horizontal and vertical velocity, angle, angular velocity, left leg ground contact, right leg ground contact) and must select one of four discrete actions at each timestep: no operation, fire left orientation engine, fire main engine, or fire right orientation engine (NOOP/LEFT/MAIN/RIGHT).

The objective is to maximize cumulative reward over an episode, with positive rewards for approaching the landing pad and negative rewards for fuel consumption and crashes. Episodes terminate upon landing and cumulative reward has been reached or in case of crash or when the configured time limit is reached. To increase task difficulty and test robustness, we enable environmental perturbations with wind power 5.0 and turbulence power 1.5, introducing stochastic forces on the lander. Success is defined as achieving cumulative reward of at least 260 on a single evaluation episode.

\subsubsection{Architecture Specification}
\bh{We employ BraiNCA as a decentralized controller for a single lunar lander agent.}
Each cell maintains a persistent state 
\bh{$\mathbf{c}_i^t \in \mathbb{R}^{C=12}$, and}
\bh{besides updating their state $\mathbf{c}_i^t\mapsto \mathbf{c}_i^{t+1}$, outputs a logit whether or not to fire, ${\boldsymbol{\ell}}_i^t=(\ell_{i,\text{noop}}^t,\ell_{i,\text{fire}}^t)$.
For a grid of $L\times M$ distributed agents to trigger one particular system level action in the context of an external environment, every cell (1) perceives the lander's state $\mathbf{o}_t$, and (2) feeds its binary activation $\ell_i^t=\arg\max(\boldsymbol{\ell}_i^t)\in\{0, 1\}$ into a majority-vote mechanism:
we compartmentalize the grid into \textit{four} zones, associate every zone with one of the \textit{four} actions (NOOP/LEFT/MAIN/RIGHT), and select the system-level action $\mathbf{u}^t\in\mathbb{R}^4$ from cellular activations $\ell_{i,\text{fire}}^t$ weighted over the respective zones.}

\bh{Local} 
aggregation \bh{either} operates over a
Moore neighborhood $\mathcal{N}_i$\bh{, and both the grid topology and layout for long-range connections $\mathcal{L}_i$ are chosen manually, see Fig.~\ref{fig:lunar_conditions}}.
Attention scores \bh{$a_{ij}^t$ and and $b_{ij}^t$, defined by Eq.~(\ref{eq:state:attention:map})}, incorporate cell states \bh{$\mathbf{c}_i^t$} \emph{and}
\bh{cell-specific previous activations $\ell_i^{t-1}$ and normalized 2D positional encodings $\rho_i$ via the extended cell state $\mathbf{s}_i^t=[\mathbf{c}_i^t;\ell_i^{t-1}; \rho_i]\in\mathbb{R}^{S=15}$.
}
\bh{Both attention maps $f_\text{attn}^\mathcal{N}(\cdot)$ and $f_\text{attn}^\mathcal{L}(\cdot)$ are two-layer MLPs (64 hidden units, GELU activation), respectively.} 




\bh{We consider four different BraiNCA conditions: either a $3\times3$ or $5\times5$ Moore Neighborhood $\mathcal{N}_i$, with or without long-range connections $\mathcal{L}_i$, respectively; 
the case without long-range connections is again referred to as Vanilla NCA. The long-range}
conditions concatenate cell state, local aggregate, and long-range contribution into \bh{the interaction}
vector
\bh{$\mathbf{z}_i^t=[\mathbf{c}_i^t;\; \mathbf{n}_i^t;\; \mathbf{l}_i^t]$, c.f., Eq.~(\ref{eq:state:composition}), while the Vanilla NCA simply uses $\mathbf{z}_i^t= [\mathbf{c}_i^t;\; \mathbf{n}_i^t; \mathbf{l}_i^t=\mathbf{0}]$.}
%
The full per-cell RNN input\bh{, $\mathbf{m}_i^t=[\mathbf{z}_i^t; \mathbf{o}_t; \mathbf{u}_{t-1}]\in\mathbb{R}^M$,} appends observation \bh{$\mathbf{o}_t$} 
and previous global action \bh{$\mathbf{u}_{t-1}$, with $M=53$ in our case.} 

At each NCA substep, the following sequence is applied for each cell\bh{: 
First, the GRU's transient state $\mathbf{h}_i^t$ is updated according to Eq.~(\ref{eq:state:update:GRU}).
Second, the refinement output $\mathbf{r}_i^t=f_\text{refine}(\mathbf{\mathbf{h}_i^t})$ is split into two readout heads, one implementing the state update $\mathbf{c}_i^{t}\mapsto\mathbf{c}_i^{t+1}=W_{C}\,\mathbf{h}_i^t$ and the other one providing the per-cell action logit $\boldsymbol{\ell}_i^t=(\ell_{i,\text{noop}}^t,\ell_{i,\text{fire}}^t)=f_{\text{act}}(\mathbf{h}_i^t) \in \mathbb{R}^{2}$; $W_{C}$ 
}
%
%
is an \bh{$M \times C$} linear \bh{projection}
and \bh{$f_\text{act}(\cdot)$}
is a two-layer MLP 
\bh{($M \to M \to 2$, GELU)}. 



The learning architecture is identical across conditions; only spatial topology and
long-range wiring differ.


\subsubsection{Spatial action selection}

\begin{figure}[h!]
\centering
    \includegraphics[width=1\linewidth]{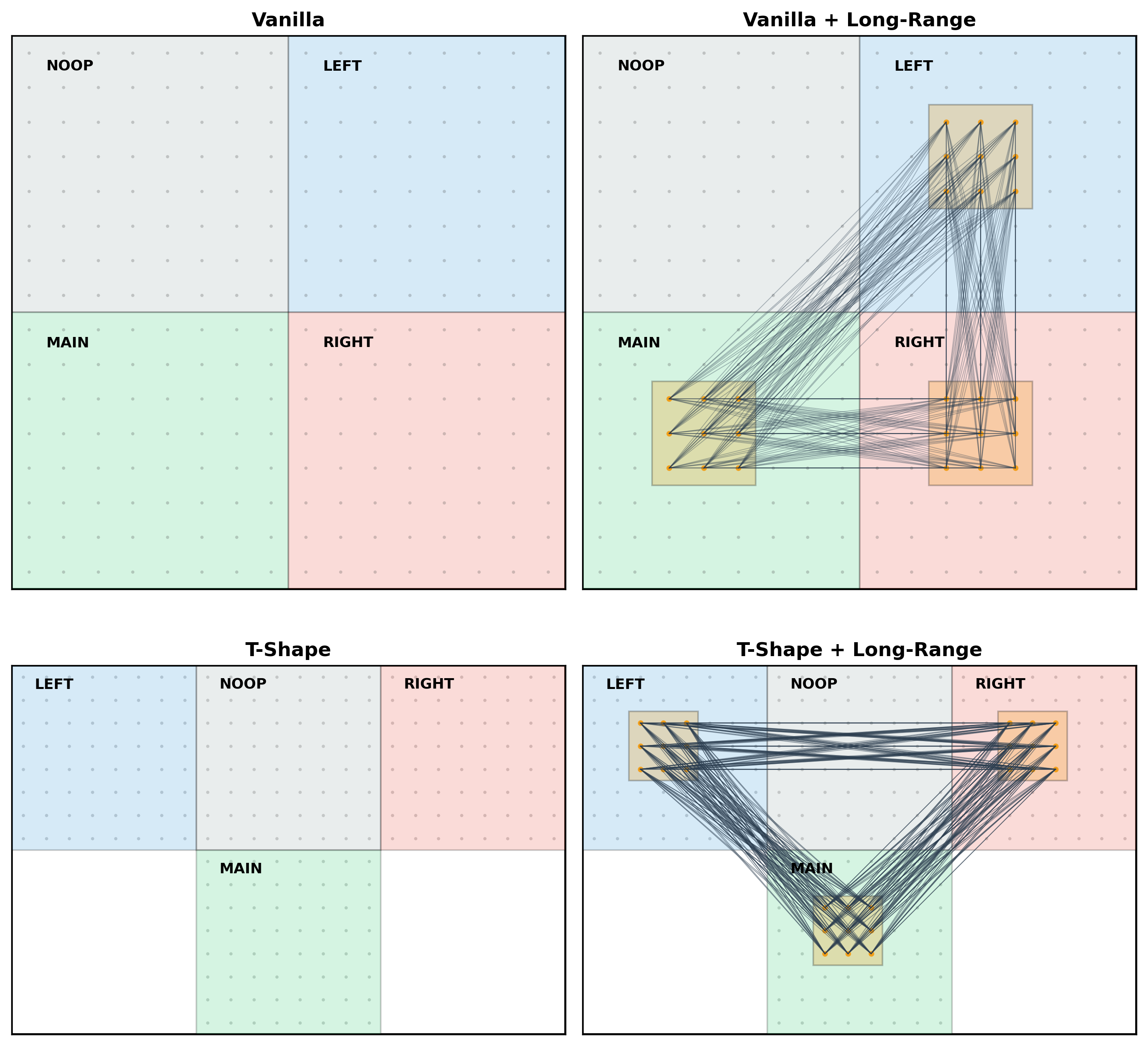}
    \caption{\textbf{LunarLander connectivity schematics for the four conditions}. Vanilla: 16$\times$16 grid with quadrant action regions (NOOP/LEFT/MAIN/RIGHT). Vanilla+LR: same grid with additional long-range links under the selected long-range architecture. T-Shape: 24$\times$16 
    \bh{grid}
    with four active 8$\times$8 zones in a T-shape. T-Shape+LR: same topology with patch-based cross-zone long-range messaging. Dots indicate individual cells \bh{in action regions (shaded blocks}.
    }
    \label{fig:lunar_conditions}
\end{figure}

Rather than producing a single global action vector, we distribute action selection spatially across the cellular grid using four dedicated action regions (NOOP, LEFT, MAIN, RIGHT) similarly to brain regions involved in motor control. In the Vanilla configuration, the 16×16 grid ($N=256$) is partitioned into four quadrants, and each action logit is read from its corresponding quadrant. In the T-shape topology, we partitioned  four distinct 8×8 quandrants too arranged in a T-shaped pattern, and each action logit is read from the corresponding zone (see Fig.~\ref{fig:lunar_conditions}). 

This spatial distribution serves multiple purposes. First, it creates a naturally interpretable mapping between spatial patterns of neural activity and behavioral outputs, analogous to motor cortex organization in biological brains where the spatial layout of neurons reflects the spatial organization of motor outputs. Second, it introduces an inductive bias that different regions of the network may specialize for different action computations, potentially facilitating learning through functional modularity. Third, the explicit spatial-action mapping with the T-shape topology may reduce the search space for effective policies by constraining the network to explore action representations that are already aligned with the task structure. 

Action selection uses a softmax-weighted mean of the fire logit within the region/quadrant. More specifically, at each environment timestep $t$, the NCA executes $n_{\text{steps}}=3$ internal updates. After these updates, each cell $i$ produces two logits:
$\boldsymbol{\ell}_i^t=(\ell_{i,\text{noop}}^t,\ell_{i,\text{fire}}^t)$.
From these two logits, we compute the cell's fire probability:
$w_i^t=\mathrm{softmax}\left((\ell_{i,\text{noop}}^t,\ell_{i,\text{fire}}^t)\right)_{\text{fire}}$.
For each action region $r\in\{\text{NOOP},\text{LEFT},\text{MAIN},\text{RIGHT}\}$,
let $\mathcal{R}_r$ be the set of cells in that region. The region logit \bh{$L_r^t$} is:
\begin{equation}
L_r^t=\frac{\sum_{i\in\mathcal{R}_r} w_i^t\,\ell_{i,\text{fire}}^t}
{\sum_{i\in\mathcal{R}_r} w_i^t+\varepsilon},\qquad \varepsilon=10^{-8}.
\end{equation}
This yields four environment-action logits
\begin{equation}
\mathbf{L}^t=[L_{\text{NOOP}}^t,L_{\text{LEFT}}^t,L_{\text{MAIN}}^t,L_{\text{RIGHT}}^t].
\end{equation}
Action selection then uses a 4-way categorical policy:
\begin{equation}
\begin{aligned}
\pi(a_t=r\mid \mathbf{o}_t) &= \mathrm{softmax}(\mathbf{L}^t)_r,\\
r &\in \{\text{NOOP},\text{LEFT},\text{MAIN},\text{RIGHT}\}.
\end{aligned}
\end{equation}

The sampled action is mapped directly to the Gym action index
$\{0,1,2,3\}\!\leftrightarrow\!\{\text{NOOP},\text{LEFT},\text{MAIN},\text{RIGHT}\}$,
executed in the environment, and then rebroadcast as one-hot previous-action memory
$\mathbf{u}_t$ for the next \bh{GRU perception cycle.}

\subsubsection{Training Procedure}

Learning employs the REINFORCE policy gradient algorithm with entropy regularization \cite{williams1992simple}. Each training episode proceeds as follows. The environment and NCA are reset to initial states (cell states initialized to zeros). At each environment step, the current observation is broadcast to all cells, the NCA executes $n_{\text{steps}}$ internal updates, action proposals are aggregated and an action is sampled from the policy distribution, and the action is executed in the environment yielding a reward and next observation. 
This continues until episode termination (landing or timeout).

The policy gradient loss is computed as:
\bh{
\begin{equation}
\mathcal{L}_{\text{policy}} = -\sum_{t=0}^{T} \log \pi(a_t \mid \mathbf{o}_t)\,\hat{R}_t - \beta\sum_{t=0}^{T}\mathcal{H}(\pi_t).
\end{equation}
}


where $\hat{R}_t$ is the discounted return from timestep $t$ (optionally normalized),
$\gamma = 0.99$, and $\beta$ is the entropy coefficient. 
\bh{During training, Gaussian noise $\xi \sim \mathcal{N}(0, 0.125)$ is added to the action logits.}
Network parameters are updated with Adam (learning rate $10^{-3}$, weight decay $10^{-4}$) and gradient clipping.
Training continues for up to 10,000 episodes with early stopping when evaluation
performance reaches or exceeds the success threshold.

\subsubsection{Long-Range Connectivity Variants}

Unlike the morphogenesis task where long-range links are random scale-free connections, the Lunar Lander long-range connections are structured by the spatial topology of the action zones. In both topologies the connections use a patch-based scheme: a 3$\times$3 patch of cells is defined at the center of each motor zone (LEFT, RIGHT, MAIN; NOOP is excluded from long-range wiring), and each patch cell receives attention-weighted signals from randomly sampled cells in the patches of the other motor zones. Each source patch cell draws 6 random targets per other zone, giving 12 long-range neighbors per connected cell.

\subsubsection{Experimental Conditions}

We evaluated four conditions with identical learning architecture and action aggregation (Vanilla NCA, Vanilla+Long-Range, T-Shape, T-Shape+Long-Range). Each condition was trained for around $1000$ independent runs.


For both tasks, we measured success rate (proportion of runs achieving performance threshold within training budget), episodes to success (number of training episodes required to first reach threshold, computed only for successful runs), and convergence variance (standard deviation of episodes-to-success across successful runs). For Lunar Lander specifically, we also recorded best evaluation reward (maximum reward achieved on evaluation episodes during training).




\section{Results}

\subsection{Long-range connections and size of neighborhood integration increase accuracy and speed of learning on the morphogenetic task}


We evaluated convergence using survival analysis on episodes-to-success, treating unsuccessful runs as right-censored at 10{,}000 episodes. Across the four morphogenetic conditions (3$\times$3 Vanilla, 3$\times$3 Long-Range, 5$\times$5 Vanilla, and 5$\times$5 Long-Range), all runs reached the success criterion (100\% success), but convergence speed differed significantly. Compared with 3$\times$3 Vanilla, 3$\times$3 Long-Range reduced RMST from 658.4 to 454.2 episodes (reduction: 204.2 episodes, 31.0\%; log-rank one-sided \(p=5.48\times10^{-6}\); RMST permutation \(p=3.20\times10^{-4}\)). Compared with 5$\times$5 Vanilla, 5$\times$5 Long-Range reduced RMST from 383.6 to 301.2 episodes (reduction: 82.4 episodes, 21.5\%; log-rank one-sided \(p=4.22\times10^{-4}\); RMST permutation \(p=4.80\times10^{-4}\)). Increasing grid size in the Vanilla setting (3$\times$3 \(\rightarrow\) 5$\times$5) reduced RMST from 658.4 to 383.6 episodes (reduction: 274.8 episodes, 41.7\%; log-rank one-sided \(p=3.28\times10^{-10}\); RMST permutation \(p=2.00\times10^{-5}\)). The best configuration (5$\times$5 Long-Range) vs baseline (3$\times$3 Vanilla) reduced RMST from 658.4 to 301.2 episodes (reduction: 357.2 episodes, 54.3\%; 2.19\(\times\) speedup; log-rank one-sided \(p=9.46\times10^{-20}\); RMST permutation \(p=2.00\times10^{-5}\)).

These results demonstrate that both architectural modifications---long-range connections and larger grid sizes---contribute independently and synergistically to improved learning efficiency (see Figure \ref{fig:results_morpho}).

\begin{figure*}[!t]
\centering
\begin{minipage}{0.74\linewidth}
    \centering
    \includegraphics[width=\linewidth]{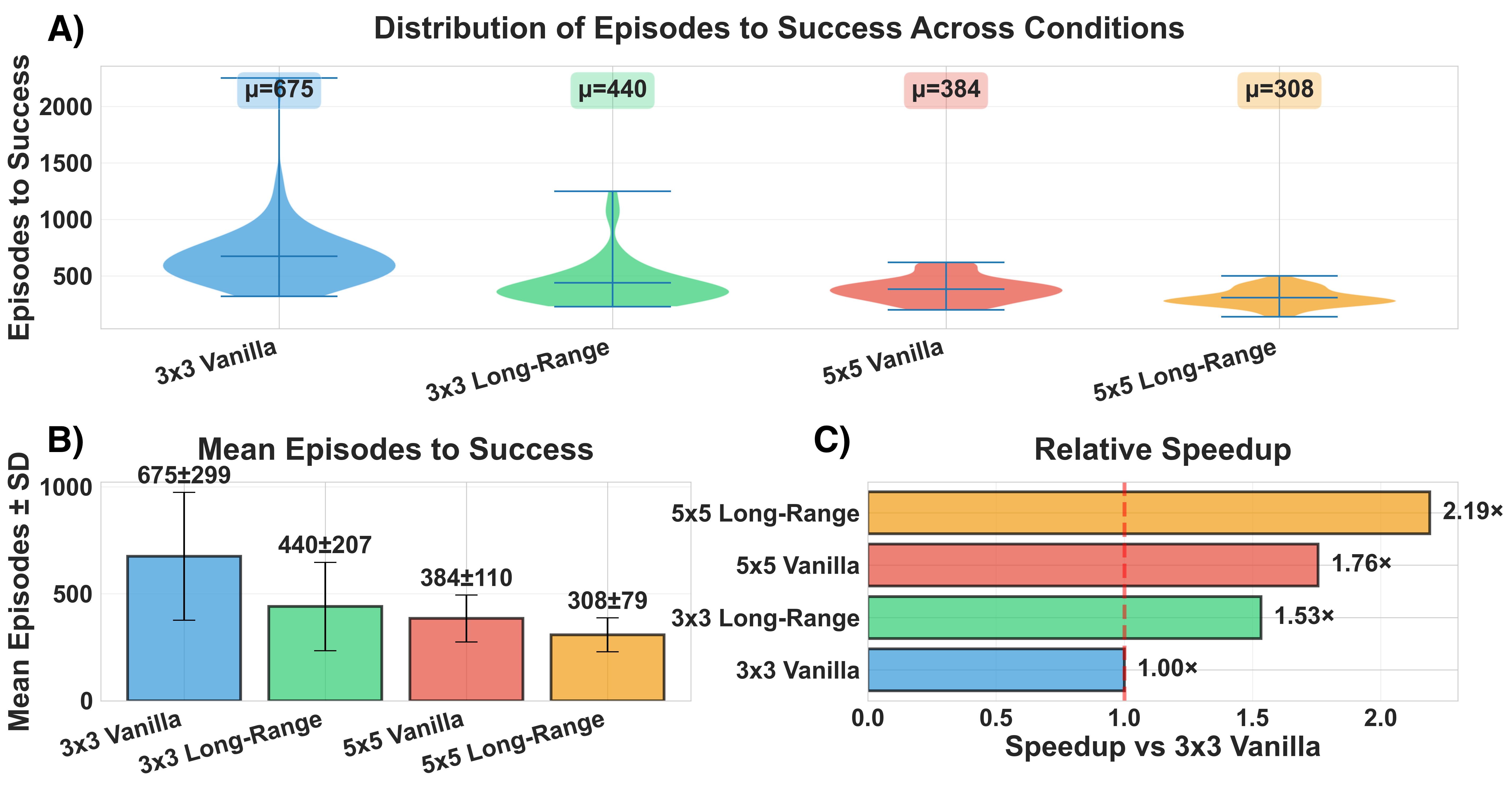}
\end{minipage}%
\hfill
\begin{minipage}{0.255\linewidth}
    \caption{\textbf{Long-range connections and size of neighborhood integration increase accuracy and speed of learning}. A)  Violin plots of the episodes to successes in the different conditions (3$\times$3 Vanilla, 3$\times$3 Long-Range, 5$\times$5 Vanilla, and 5$\times$5 Long-Range). B) Mean episodes to success in the different conditions. C) Relative speedup compared to the baseline (3$\times$3 Vanilla).}
    \label{fig:results_morpho}
\end{minipage}
\end{figure*}

\subsection{Somatotopic topology increase robustness and speed of learning} 

\begin{figure*}[!t]
\centering
\begin{minipage}{0.74\linewidth}
    \centering
    \includegraphics[width=\linewidth]{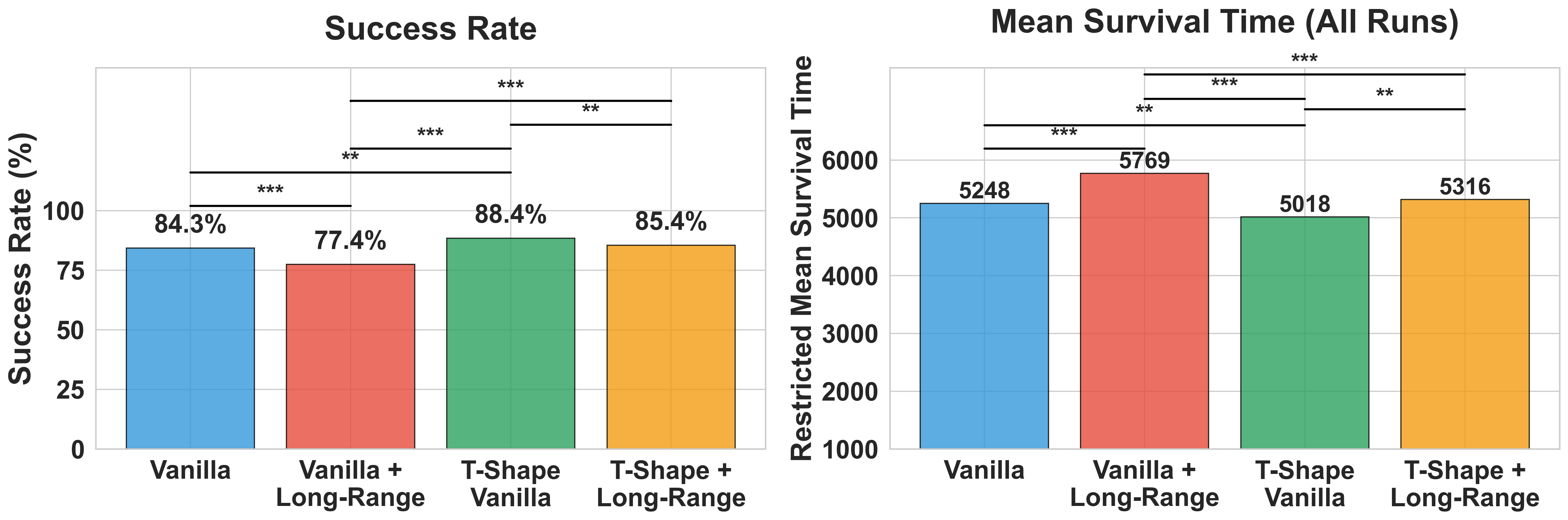}
\end{minipage}%
\hfill
\begin{minipage}{0.255\linewidth}
    \caption{\textbf{Somatotopy increase robustness and speed of learning to the task}. Left. Success rate in the different conditions (Vanilla, Long-Range with quadrant-patch connections, and T-Shape topology). Right. Mean survival time in the different conditions. \bh{*:\,$p < 0.05$, {**:\,$p < 0.01$}, ***:\,$p < 0.001$}} 
    \label{fig:lunar}
\end{minipage}
\end{figure*}

We evaluated four NCA architectures on the LunarLander task: (1) Vanilla (baseline, $n=1300$), (2) Long-Range with quadrant-patch connections ($n=1250$), (3) T-Shape Vanilla with T-shaped topology but no long-range connections ($n=1279$), and (4) T-Shape + Long-Range, combining the T-shaped topology with quadrant-patch long-range connections ($n=1344$) (see Figure \ref{fig:lunar}).

Across conditions, success rates were $84.3\%$ for Vanilla, $77.4\%$ for Long-Range, $88.4\%$ for T-Shape Vanilla, and $85.4\%$ for T-Shape + Long-Range. Vanilla was significantly more robust than Long-Range (Fisher's exact one-sided test, $p=3.90 \times 10^{-6}$). T-Shape Vanilla showed a significantly higher success rate than Vanilla (Fisher's exact one-sided test, $p=0.00137$) and than Long-Range ($p=1.01 \times 10^{-13}$). T-Shape + Long-Range also achieved a significantly higher success rate than Long-Range ($p=1.05 \times 10^{-7}$), whereas it did not significantly improve robustness relative to Vanilla. Survival analysis over all runs, with unsuccessful runs right-censored at $10{,}000$ episodes, further showed that Vanilla converged faster than Long-Range (log-rank one-sided $p=1.35 \times 10^{-6}$; RMST permutation $p \approx 0$). T-Shape Vanilla also converged faster than Vanilla (log-rank one-sided $p=0.00666$; RMST permutation $p=0.0251$) and than Long-Range (log-rank one-sided $p=3.72 \times 10^{-13}$; RMST permutation $p=2.00 \times 10^{-5}$). T-Shape + Long-Range also converged faster than Long-Range (log-rank one-sided $p=1.71 \times 10^{-6}$; RMST permutation $p=6.00 \times 10^{-5}$), but did not significantly differ from Vanilla. The corresponding restricted mean survival times were $5248.0$ episodes for Vanilla, $5768.6$ for Long-Range, $5018.1$ for T-Shape Vanilla, and $5316.3$ for T-Shape + Long-Range, indicating that Vanilla learned $520.6$ episodes faster than Long-Range, while T-Shape Vanilla reduced mean time-to-success by $229.9$ episodes relative to Vanilla and by $750.5$ episodes relative to Long-Range.


The higher success rates and speed of learning in the  T-Shape conditions  highlight the benefits of the somatotopic regionalization and topology. In all conditions, action selection is spatially aggregated, but T-Shape explicitly partitions the grid into four zones (LEFT, RIGHT, MAIN, NOOP), mirroring the spatial-functinal relationshipp with the lunar lander. This is similar to the somatotopic organization in motor cortex like spatial-functional relationships \cite{schieber2001constraints}. This spatial specialization appears to yield more robust task performance and faster speed of learning relative to standard Vanilla, suggesting that NCA regionalization and new topologies contribute to improved learning. Interestingly on this task, the long-range connections present a negative effect on robustness and speed of learning compared to the Vanilla condition. One possible reason may be the simplicity of the patch organization of the long-range connection and future studies will explore the learning of the optimal connectivity for a BraiNCA.

We can conclude that new architectural modifications, e.g. topology with a spatial/functional organization improve robustness and learning speed.

\section{Discussion}

In this paper, we introduce BraiNCA, an NCA model equipped with an attention layer, long-range connections, and a novel topology design. BraiNCA replaces purely local aggregation with attention-weighted neighborhood integration, allowing each node to dynamically select and gate information from its neighbors and long-range ones based on contextual relevance. Beyond this, we explicitly model long-range interactions by augmenting the underlying connectivity (e.g., through multi-hop edges or dedicated long-range connections), enabling information flow that extends beyond one-hop (Moore) neighborhoods as in Vanilla NCAs. And to further capture spatial–functional relationships reminiscent of neural architectures like in somatotopy, we introduce a new T-shaped topology that mimics structured, spatially grounded connectivity patterns for the lunar lander task. This design departs from the standard grid topology traditionally used in NCAs, offering a more expressive substrate to emulate heterogeneous and functionally specialized wiring, such as those observed in biological systems. We benchmark the BraiNCA architecture on two tasks, morphogenesis and motor control experiments. 
BraiNCAs learn faster and show improved robustness compared to Vanilla NCAs in the two tasks.


The morphogenetic experiment suggests that long-range information increases survival in development and morphogenesis. Cells benefited from long-range signalling in our framework via two mechanisms, long-range connections and increase in neighborhood integration 
{\color{black} -- both} found in biology{\color{black}:}
{\color{black} (i) Tunneling} nanotubes (TNTs), for example, {\color{black} can form}
 cytoplasmic bridges for long-range, directed communication and have been reported to reach distances up to ~300 µm \cite{ariazi2017tunneling}.
{\color{black}(ii) Gap}-junction–coupled cell networks provide a natural substrate for multi-hop propagation of ionic and small-molecule signals, allowing 
{\color{black} locally sensed perturbations}
to influence a larger neighborhood or distant cells. Bioelectrical networks may allow long-range communication during morphogenesis \cite{levin2014endogenous, levin2021bioelectric}.{\color{black}

}
Mechanistically, long-range information can act like a supervisory layer: it stabilizes attractor states in gene regulatory networks, coordinates timing across fields of cells, and aligns growth with target morphology. These observations motivate a view in which developmental control is implemented not only by diffusible morphogen gradients or local gradients, but also by intercellular communication architectures that can transmit state information across multiple cells/neighbors and across many different tissues. In particular, this kind of long-range coordination may be especially critical during periods of rapid proliferation or remodeling, when local noise is high and the cost of developmental error rises.


The current model imposes a fixed long-range (and multi-hop) topology, and thus puts an inductive bias on the BraiNCA model. Future work will be dedicated to employ developmental principles to ``grow'' a per-cell long-range affinity matrix across the entire multicellular grid alongside the typical cell-state updates that govern NCA dynamics. This will enable a joint development of morphology and long-range connectivity akin to a virtual organism with a plastic digital nervous system. More broadly, BraiNCA provides brain-inspired NCA formulation that preserves the decentralized “local update” principle while better reflecting non-local connectivity patterns, making it a promising substrate for studying collective computation under biologically-realistic network structure and evolving cognitive substrates.

\newpage
\section*{Acknowledgements}

We thank members of the Levin Lab for helpful discussions. 
We gratefully acknowledge support for this work provided through a sponsored research agreement with Astonishing Labs and from the Templeton World Charity Foundation, Inc. (Grant ID: TWCF-2021-20606). The opinions expressed in this publication are those of the authors and do not necessarily reflect the views of the funding agencies.
\bibliography{references}
\bibliographystyle{apalike}


\end{document}